\documentclass[a4paper,twoside]{article}

\usepackage{epsfig}
\usepackage{subcaption}
\usepackage{calc}
\usepackage{amssymb}
\usepackage{amstext}
\usepackage{amsmath}
\usepackage{amsthm}
\usepackage{pslatex}
\usepackage{multicol} 
\usepackage{url} 
\usepackage{booktabs}
\usepackage{multirow}
\usepackage{tabularx} 
\usepackage{apalike}
\usepackage{algorithm2e}

\usepackage[table]{xcolor}

\usepackage{listings}
\usepackage[bottom]{footmisc}
\usepackage{SCITEPRESS}     

\begin{document}

\title{Analysis of LLM Performance on AWS Bedrock: Receipt-item Categorisation Case Study}

\author{
\authorname{Gabby Sanchez, Sneha Oommen, Cassandra T. Britto, Di Wang, Jung-De Chiou, Maria Spichkova} 
\affiliation{  RMIT University, Melbourne, Australia}
}
 

\abstract{This paper presents a systematic, cost-aware evaluation of large language models (LLMs) for receipt-item categorisation within a production-oriented classification framework. We compare four instruction-tuned models available through AWS Bedrock: Claude 3.7 Sonnet, Claude 4 Sonnet, Mixtral 8x7B Instruct, and Mistral 7B Instruct. The aim of the study was (1) to assess performance across accuracy, response stability, and token-level cost, and (2) to investigate what prompting methods, zero-shot or few-shot, are
especially appropriate both in terms of accuracy and in terms of incurred costs.  
Results of our experiments demonstrated that Claude 3.7 Sonnet achieves the most favourable balance between classification accuracy and cost efficiency.\\
\emph{Preprint. Accepted to the 19th International Conference on Evaluation of Novel Approaches to Software
Engineering (ENASE 2026). Final version to be published by SCITEPRESS, http://www.scitepress.org}}

\onecolumn \maketitle \normalsize \setcounter{footnote}{0} \vfill
 
\section{\uppercase{Introduction}}

Accurate text categorisation is crucial in various application areas, including legal and financial documents as well as social media. For financial documents like quotes and receipts, precise categorisation is essential in order to derive meaningful financial insights and ensure a reliable user experience. 
With the rise of the capacity of large language models (LLMs), many studies investigated their applicability for the text categorisation tasks, see, for example,  \cite{wan2024tnt} and \cite{wang2025survey}. 
In this study, we examine the limitations of existing LLMs for the task of expense item categorisation using a curated dataset of diverse receipts. We also analyse what prompting methods, zero-shot or few-shot~\cite{liu2023pre}, are most appropriate both in terms of accuracy and incurred costs. 
Our goal is to establish a strong foundation for further analysis of receipt content at scale. 
We aim to answer the following research questions:
\\
\emph{RQ1. How effective can LLMs be for receipt items categorisation?}
\\
\emph{RQ2. What LLM might be especially appropriate for the task of receipt items categorisation?}
\\
\emph{RQ3. What might be the most appropriate prompting method for the task of expense-items categorisation? } 

To select a suitable classification model, we conducted a structured evaluation of several foundational models available through AWS Bedrock. The goal was to determine whether these models could reliably interpret receipt data already available in text form and classify items according to the application’s predefined categories. We compared their accuracy, consistency, runtime performance, and cost efficiency using a shared dataset of real receipts.
Based on these findings, we refined the category definitions and prompt design to improve classification accuracy and interpretability and tested whether few-shot examples would provide additional benefit. This iterative approach ensured that the final configuration achieved both high accuracy and cost-effective performance.  

AWS Bedrock provides access to multiple foundational models through a unified \emph{InvokeModel API} that
accepts a prompt and returns a model-generated response~\cite{AWSBedrock}, \cite{AWSInvokeModel}. This platform allows developers to test different language models using a consistent interface and security framework, making it suitable for controlled evaluations.

\emph{Contributions:}  We conducted a comparative analysis of four models hosted on Bedrock: two proprietary models from Anthropic, Claude Sonnet 3.7  and Claude Sonnet 4, and two open-weight models from Mistral, Mixtral 8x7B Instruct and Mistral 7B Instruct.
All models were evaluated using a curated dataset of receipt line items mapped to a structured category taxonomy, with identical schema-first prompts to ensure a fair comparison.   
Claude Sonnet models have been selected as they demonstrated solid results in text classification in related areas. Mistral models have been selected for inclusion in the study as they are open-weight.

\section{\uppercase{Related Work}}
\label{sec:related}

\subsection{LLM performance in text classification and annotation}

There are numerous studies comparing the performance of various LLMs. However, to our best knowledge, none of the studies so far have compared the performance of the Claude and Mistral models in the categorisation of receipt items. In this section, we discuss studies comparing LLM performance across related areas. 
For example, the survey presented in \cite{wang2025survey} provides an overview of LLMs applied to text classification. 
The study presented in \cite{wei2023empirical} introduced experiments on  text classification in legal documents conducted with a pre-trained DistilBERT model.

A number of studies also analysed the capacity of LLMs for text annotation. For example, the study presented in \cite{nasution2024chatgpt} focused on ChatGPT-4, while the work presented in~\cite{rouzegar2024enhancing}   employed the GPT-3.5. 
The study introduced in \cite{pangakis2024knowledge} compared GPT-4 and Mistral-7B performance.  
A set of best practices for text annotation with LLMs was presented in~\cite{tornberg2024best}. 

The study presented in \cite{jauhiainen2024evaluating} compared the effectiveness of LLMs ChatGPT-3.5, ChatGPT-4, Claude-3, and Mistral-Large in evaluating university students' open-ended responses. The analysis has been conducted using the models available in 2024, leading to the conclusion that the most accurate results were obtained with Claude-3. A similar work has been presented in \cite{lobo2025automatic}, where the authors compared the effectiveness of Gemini 2.0 flash, GPT-4o, CSonnet 3.7, and Mistral Large in evaluating essays written in Brazilian Portuguese by primary school students. The authors concluded that Claude Sonnet 3.7 was especially aligned with human markers, suggesting its potential for automated marking. These findings correlate with the results of our study.

The study presented in \cite{spichkova2025analysis}
introduced a performance analysis of AWS Rekognition and Azure Custom Vision for the recognition of parking signs, where AWS Rekognition demonstrated solid results. 
In our previous studies~\cite{spichkova2019easy,spichkova2019comparison}, 
we investigated the performance of Google Cloud Vision and AWS Rekognition for the automation of the meter reading process for the standard (non-smart) meters using computer vision techniques. The study demonstrated that AWS Rekognition provides better results for reading data from non-smart meters. However, over the last years, many other studies have been conducted based on AWS technologies. In what follows, we provide a brief overview of them.
 
 \subsection{Research embedded in T\&L process}

Enhancing students' exposure to research activities is a crucial part of the curriculum. Nevertheless, there is no agreement on the best way to achieve this. 
Holz at al.~\cite{holz2006research} proposed a framework for teaching Computing
Research Methods (CRM). 
Weinman et al.~\cite{weinman2015teaching}  providing summer research experiences for undergraduate students. 
Fernandez~\cite{fernandez2024generalizing} described an initiative at the University of Texas to offer course-based research experiences for undergraduate students, where they work on research tasks individually. 
Host~\cite{host2002introducing} proposed incorporating empirical software engineering methods into project-based courses as part of the study curriculum, which aligns closely with the spirit of our initiative, proposed at the RMIT University~\cite{spichkova2017autonomous,simic2016enhancing}.

To encourage the curiosity of Bachelor's and Master's students about research in Software Engineering, we suggest incorporating research and analysis components into projects as an optional bonus task. 
For instance, Sun et al.~\cite{sun2018software} and Clunne-Kiely et al.~\cite{clunne2017modelling} examined the HCI aspects of an autonomous humanoid robot.
Spichkova et al.~\cite{spichkova2020gosecure}  introduced an automated solution for scanning security vulnerabilities in Google Cloud Platform projects. 
Several other projects focused on visualisation aspects, e.g.,   visualisation of AWS service usage~\cite{george2020usage},  
visualisation of the CIS benchmark scanning results~\cite{zhao2025visualisation}, and   
visualisation and analysis of data collected from vertical transport facilities~\cite{christianto2017software}. 

 \begin{figure*}[ht!]
     \centering
     \includegraphics[width=0.8\linewidth]{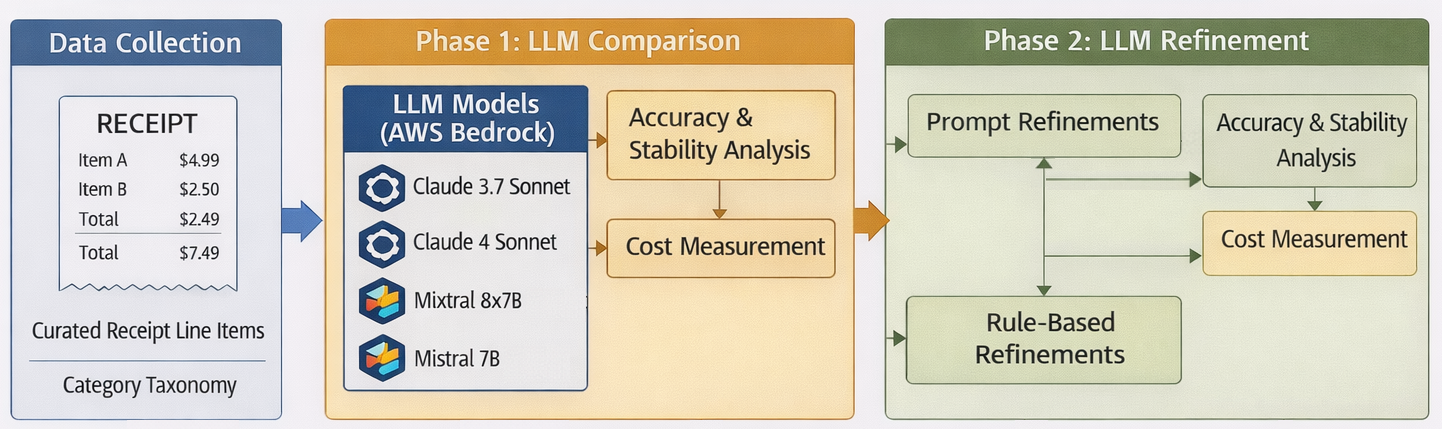}
     \caption{Methodology of the conducted study (Disclaimer: The diagram was elaborated in collaboration with ChatGPT-5.2, based on a detailed textual description of the required diagram)}
     \label{fig:methodology}
 \end{figure*}

\section{\uppercase{Case study: Methodology}}
\label{sec:methodology}

The overall methodology of the conducted case study is presented in Figure~\ref{fig:methodology}. 
After collecting the data and creating a curated data set, the comparative study was conducted in two phases. 

The primary objective of Phase 1 was to identify the most suitable foundational model available through AWS Bedrock for integration into a framework for automated collection and analysis of shopping receipts. We conducted a comparative evaluation of four models: Claude 3.7 Sonnet, Claude 4 Sonnet, Mixtral 8x7B Instruct, and Mistral 7B Instruct.  All models were evaluated using a curated dataset of receipt line items mapped to a structured category taxonomy, with identical schema-first prompts to ensure a fair comparison. 
In addition to model selection, we aimed to establish a baseline understanding of how different model families interpret receipt item text and adhere to schema-based output requirements. The results from this stage were used to identify strengths and limitations of each approach, providing a foundation for refinements later in the
evaluation, during which category definitions and prompt design were systematically improved.

The objectives of Phase 2 were to improve and validate the model selected in Phase 1 by testing
prompt-level refinements and selecting a production configuration with the best accuracy-to-cost balance. 
We compared variants that update the predefined categories, introduce rule-based disambiguation, and add a few-shot examples. 
Each variant is evaluated on the same dataset using the methods applied in Phase 1, with results
reported for overall and balanced accuracy, precision, recall, F1 score, runtime, and estimated cost per call based on token usage. We also run a lenient check that accepts plausible alternative categories to quantify inherent
label ambiguity. 
The outcome of Phase 2 is a single recommended configuration and clear guidance on trade-offs for production.

\subsection{Data Collection}
The evaluation used a single CSV exported from the AWS Textract Evaluation as its primary data, see~\cite{dataSet}. 
Each row represented one receipt, containing machine-extracted text blocks, the detected vendor, and the total amount. The dataset comprised  
389 expense items.  
The data set served as input for all model evaluation scripts and was used consistently across both Phases 1 and 2 to ensure comparable results.

Because the Textract input dataset was unevenly distributed, some categories, such as `Pantry \& Snacks' and `Eating Out, appeared more frequently than others, like `Pets' or `Subscriptions \& Digital Services'.
The data set didn't include items belonging to represent `Baby \& Maternity', `Entertainment', or `Transport \& Fuel', leaving those
categories without ground-truth examples at this phase of the evaluation. 
Balanced accuracy and per-category reporting were therefore used to offset these imbalances. 
Ground truth labels were manually assigned to each
receipt and item using the Phase 1 baseline category set. 
We applied a labelling guide to clarify boundaries between similar categories, such as `Eating Out' versus `Pantry \& Snacks' and `Coffee \& Tea' versus `Beverages'.
When items fit multiple plausible categories, we applied a strict interpretation and recorded only one target label.

Ground-truth labels were manually prepared by the first and second authors. Each receipt item was inspected and assigned a single definitive category using the Phase 1 baseline category set. The resulting \texttt{ground\_truth\_category} column
was then merged into the output from the Step 1 script before evaluation by the Step 2 evaluator. This process
ensured that correctness scoring was based on verified human annotations.

\subsection{Methods applied in Phase 1}

Each model was treated as a black box,   
where performance was assessed using the same data set and consistent prompt structures, with evaluation criteria covering classification accuracy, consistency, hallucination rate, runtime, and token usage as a proxy for cost. 
Although the InvokeModel API is shared across providers, prompt formatting differs by family:
Claude models separate system and user messages, whereas Mixtral and Mistral use a single combined text input~\cite{AWSModels}.

In addition to model selection, we aimed to establish a baseline understanding of how different model families interpret receipt item text and adhere to schema-based output requirements. The results from this stage were used to identify strengths and limitations of each approach, providing a foundation for refinements later in the evaluation, during which category definitions and prompt design were systematically improved.

The study was designed to identify an approach that would produce results with high accuracy but at minimal cost. For these reasons, the Bedrock InvokeModel API was used, with all processing performed locally to avoid charges from services such as S3, DynamoDB, Lambda, or Textract. 
Model invocations were executed by the model evaluator script,  
which iterated over every receipt, invoked each model via the Bedrock InvokeModel API, and logged detailed outputs to CSV files. 
The script captured timing data, token counts (when reported), and validation errors for
each call. 
Concurrency was kept to a minimum to avoid rate limiting and maintain consistent runtime conditions.
Correctness scoring and metric computation were performed by the ground truth evaluator script, 
which compared each model’s predictions with the
manually integrated ground-truth column.

Prompts were concise and free of few-shot examples to limit input tokens, see~\cite{dataSet}, and the required JSON-array output kept responses short. All input and output were handled as local CSV files, and successful results were cached to prevent unnecessary re-invocations. These measures ensured reproducibility while keeping per-receipt costs low.

\subsubsection{Prompt Engineering}
Prompts were designed to maintain identical task instructions and schema requirements across all model families while adapting to provider-specific input formats. 
For Claude models, a two-part structure was used:
The system message defined the classification task, schema, and constraints, and the user message contained the serialised receipt text (vendor, total, and item list). 
For Mixtral models, a single combined
prompt embedded both the task instructions and the receipt data.
To ensure that the models remained tightly bound to valid outputs, each prompt explicitly included a predefined list of categories that are used by the proposed framework. 
By constraining the model to select from a fixed
set, rather than generating free-form labels, we reduced ambiguity and improved consistency across runs. This design choice also ensured alignment with the application’s reporting logic, which requires that each receipt item belong to exactly one recognised category.
This schema-first design improved parseability, minimised hallucinations, and reduced token usage through concise wording, following AWS Bedrock’s official guidance on concise, schema-driven prompt design~\cite{AWSprompt1}.

Each prompt required a JSON array of category labels, each corresponding to the number of detected items,
with no commentary or extra text. Simple heuristics, such as classifying cafes and restaurants as `Eating Out' or `Coffee \& Tea', were included to improve consistency.
The complete set of prompt templates, including schema definitions, heuristics, and model-specific variations,
is provided in supplementary materials~\cite{dataSet}.

\subsubsection{Predefined Category Set}
The predefined category set used in the prompts was created before any model evaluation. 
It was elaborated collaboratively by the authors during numerous meetings. To ensure relevance to Australian consumer spending, the team also reviewed how local financial and budgeting applications (such as the Commonwealth Bank mobile app) group transactions. 
The final list was designed to cover common household, personal, and service-related expenses while maintaining clear boundaries between similar categories.
The full list of 26 categories and their identifiers is shown in Table~\ref{tab:expense_categories}. The same list is hard-coded in the prompt templates, ensuring that all model runs referenced identical labels and schema definitions.

\begin{table}[ht]
\centering
\caption{Predefined Category Set used in Phase 1}
\label{tab:expense_categories}
\renewcommand{\arraystretch}{1}
\scriptsize{
\begin{tabular}{cll}
\toprule
\textbf{\#} & \textbf{Category Name} & \textbf{Identifier} \\
\midrule
1  & Fresh Produce & \texttt{fresh\_produce} \\
2  & Meat \& Seafood & \texttt{meat\_seafood} \\
3  & Dairy \& Eggs & \texttt{dairy\_eggs} \\
4  & Pantry \& Snacks & \texttt{pantry\_snacks} \\
5  & Bakery & \texttt{bakery} \\
6  & Coffee \& Tea & \texttt{coffee\_tea} \\
7  & Beverages & \texttt{beverages} \\
8  & Alcohol & \texttt{alcohol} \\
9  & Eating Out & \texttt{eating\_out} \\
10 & Health \& Medicine & \texttt{health\_medicine} \\
11 & Personal Care \& Beauty & \texttt{personal\_care\_beauty} \\
12 & Home \& Cleaning & \texttt{home\_cleaning} \\
13 & Baby \& Maternity & \texttt{baby\_maternity} \\
14 & Pets & \texttt{pets} \\
15 & Clothing \& Footwear & \texttt{clothing\_footwear} \\
16 & Electronics \& Tech & \texttt{electronics\_tech} \\
17 & Stationery \& Office & \texttt{stationery\_office} \\
18 & Sports \& Fitness & \texttt{sports\_fitness} \\
19 & Gifts \& Occasions & \texttt{gifts\_occasions} \\
20 & Entertainment & \texttt{entertainment} \\
21 & Subscriptions \& Digital Services & \texttt{subscriptions\_digital} \\
22 & Professional Services & \texttt{professional\_services} \\
23 & Utilities \& Bills & \texttt{utilities\_bills} \\
24 & Transport \& Fuel & \texttt{transport\_fuel} \\
25 & Travel \& Holidays & \texttt{travel\_holidays} \\
26 & Other & \texttt{other} \\
\bottomrule
\end{tabular}
}
\end{table}

 The evaluation used receipts collected from Australian vendors. The dataset was manually gathered
by the authors. 65\% of the receipts   were physical receipts photographed using three different smartphones: Samsung Galaxy A52 (Android 14),
 Google Pixel 7 Pro (Android 16), and iPhone 13 Pro (iOS 18.6.2). The remaining 35\% of the receipts  were electronic receipts obtained as screenshots of digital copies. 
 The study was limited to receipts in JPG and PNG formats, taken with a camera or via screenshots from mobile phone apps (both Android and iOS). Thus, all  receipts were stored as image files, and no PDF-based receipts were included.

\subsubsection{Data Analysis and Evaluation}

Correctness was determined automatically by comparing each model’s predicted category with the manually
assigned ground-truth label.  
A prediction was counted as correct only when it exactly matched the corresponding ground-truth category for that item.
Model performance was evaluated using several standard classification metrics:
\begin{itemize}
    \item \emph{Accuracy} (overall accuracy) represents the proportion of correctly predicted items across the entire dataset.
    \item \emph{Balanced accuracy}  is the average of per-category accuracies, giving equal weight to each category regardless of how many items it contains; this helps counter the imbalance in category frequencies.
    \item  \emph{Precision} measures the proportion of items a model correctly labels as a given category.
    \item \emph{Recall} measures the proportion of items from a given category that were successfully identified.
    \item The \emph{F1 score} is the harmonic mean of precision and recall.
\end{itemize}

Metrics were calculated using both micro averaging (equal weighting of all predictions) and macro averaging (averaging scores per category). Together, these measures provide a balanced view of classification quality across frequent and infrequent categories.

All metrics were computed across all 389 items, along with per-category accuracy tables and
support counts.  
Ground-truth categories were then manually added to the evaluation results file by the first and the second authors. 
This labelled file was subsequently used by the evaluator script 
to automatically compute all performance metrics. 
All performance metrics were calculated using the complete dataset, ensuring accuracy, consistency, and reproducibility of the results. 

\subsection{Methods applied in Phase 2}
\label{sec:methods_phase1}

Phase 2 followed the same evaluation process as Phase 1, focusing exclusively on prompt-level refinements to the model selected in Phase 1's results, i.e., the Claude 3.7 Sonnet model. 
The primary goal was to determine whether structured prompt updates could improve classification accuracy without compromising cost efficiency or response consistency. 
Unless otherwise stated, all configuration parameters, dataset inputs, and evaluation metrics remained identical to Phase~1 to ensure comparability. 
The subsections below describe the elements unique to Phase~2, including updates to the predefined category set, prompt variants, evaluation procedure, data analysis, and cost considerations.

\subsubsection{Prompt Engineering}
The same cost-control measures from Phase 1 were retained: all experiments were executed locally using only the Bedrock InvokeModel API, with results cached to avoid redundant inferences. The main cost difference in Phase 2 arose from longer prompt text and, in the few-shot variant, embedded examples, both of which increased input token counts. 
Output length remained stable across variants. 
As a result, input token usage became the
primary cost driver, with the most tuned variant roughly doubling the per-call cost compared to the baseline. 
These findings were integrated into the accuracy-to-cost comparison to support model selection for deployment.

Phase 2 tested four prompt configurations derived from the Claude 3.7 Sonnet baseline, see Table~\ref{tab:phase2-variants}:
\begin{enumerate}
    \item \textbf{Variant 1}:
    The baseline variant (applied in Phase 1) using the original predefined category set, and no rule-based adjustments or examples.
    \item \textbf{Variant 2}:
Revised category set with the same
schema, but no additional heuristics or examples.
    \item \textbf{Variant 3}:
Revised category set with added explicit classification rules and disambiguation heuristics to minimise overlap (e.g., \emph{`If an item is refrigerated or perishable, classify it
under dairy\_eggs\_fridge rather than pantry\_snacks'}).
    \item \textbf{Variant 4}:
    Revised category set with added explicit classification rules and disambiguation heuristics to minimise overlap, as well as added several examples to illustrate the expected format and logic.
\end{enumerate}

\begin{table}[ht]
\centering
\caption{Comparison of the applied configurations}
\label{tab:phase2-variants}
\renewcommand{\arraystretch}{1}
\scriptsize{
\begin{tabular}{l c c c  }
\toprule
\textbf{Variant} & \textbf{Category set} & \textbf{Extra rules} & \textbf{Zero/few shot}   \\
\midrule
Variant 1 & Baseline & -- & Zero-shot     \\
Variant 2 & Refined  & -- & Zero-shot   \\
Variant 3 & Refined  & \checkmark & Zero-shot     \\
Variant 4 & Refined  & \checkmark &  Few-shot   \\ 
\bottomrule
\end{tabular}
}
\end{table}

Each prompt retained the schema-first design from Phase 1, requiring output as a JSON array whose length matched the number of detected items. 
Wording was kept concise to balance clarity and token efficiency. The classification rules covered the full category set and included boundaries such as:
\begin{itemize}
    \item 
    Distinguishing refrigerated or perishable items from shelf-stable goods (\emph{dairy\_eggs\_fridge} vs. \emph{pantry\_snacks});
    \item  
    Separating drinkable liquids from pantry items (\emph{drinks} vs. \emph{coffee\_tea});
    \item 
    Prioritising context clues such as vendor type or packaging keywords (\emph{cafe} or \emph{takeaway}  to \emph{eating\_out}).
    \item
    Handling ambiguous personal or household items (\emph{personal\_care\_beauty} vs.  \emph{cleaning\_maintenance}).
\end{itemize}

These rules were formulated to guide consistent boundary handling rather than promote memorisation,
following AWS Bedrock’s official prompt engineering recommendations~\cite{AWSprompt1}. 

\subsubsection{Refined Category Set}

The baseline category list from Phase 1 was revised following a review of ambiguous or overlapping labels
observed during data annotation and model evaluation. Several receipt items, such as frozen dumplings, hummus, deli meats, and iced coffee drinks, highlighted weaknesses in the earlier set, particularly in distinguishing between perishable goods and cold-storage items. To address this, the updated list introduced finer-grained categories that clarified boundaries between similar domains, see Table~\ref{tab:phase2_categories}. 

\begin{table}[ht]
\centering
\caption{Updated Predefined Category Set used in Phase 2’s Model Evaluation}
\label{tab:phase2_categories}
\scriptsize{
\renewcommand{\arraystretch}{1}
\begin{tabular}{cll}
\toprule
\textbf{\#} & \textbf{Category Name} & \textbf{Identifier} \\
\midrule
1  & Fruits \& Vegetables & fruits\_vegetables \\
2  & Meat \& Seafood \& Deli & meat\_seafood\_deli \\
3  & Dairy \& Eggs \& Fridge & dairy\_eggs\_fridge \\
4  & Frozen & frozen \\
5  & Pantry \& Snacks & pantry\_snacks \\
6  & Bakery & bakery \\
7  & Coffee \& Tea & coffee\_tea \\
8  & Drinks & drinks \\
9  & Liquor & liquor \\
10 & Eating Out & eating\_out \\
11 & Health \& Medicine & health\_medicine \\
12 & Personal Care \& Beauty & personal\_care\_beauty \\
13 & Cleaning \& Maintenance & cleaning\_maintenance \\
14 & Baby \& Maternity & baby\_maternity \\
15 & Pets & pets \\
16 & Clothing \& Footwear & clothing\_footwear \\
17 & Electronics \& Tech & electronics\_tech \\
18 & Home \& Lifestyle & home\_lifestyle \\
19 & Sports \& Fitness & sports\_fitness \\
20 & Gifts \& Occasions & gifts\_occasions \\
21 & Entertainment & entertainment \\
22 & Subscriptions \& Digital Services & subscriptions\_digital \\
23 & Professional Services & professional\_services \\
24 & Utilities \& Bills & utilities\_bills \\
25 & Transport \& Fuel & transport\_fuel \\
26 & Travel \& Holidays & travel\_holidays \\
27 & Other & other \\
\bottomrule
\end{tabular}
}
\end{table}

\subsubsection{Data Analysis and Evaluation}

The same dataset of  389 expense items and the orchestration process from Phase 1 were reused to
ensure consistency. All four prompt variants were tested using identical invocation settings via AWS Bedrock’s InvokeModel API. 
Ground-truth categories were manually verified and updated by the first and the second authors  
prior to evaluation.  

Evaluation continued under two modes:
\begin{itemize}
    \item 
    \textbf{Strict evaluation}, identical to Phase 1, where a prediction was considered correct only if it exactly matched the ground-truth category.
    \item 
    \textbf{Lenient evaluation}, newly introduced in Phase 2, where a prediction was counted as correct if it matched either the primary category or a valid alternative category, accounting for naturally ambiguous receipt items (e.g., `iced coffee' could reasonably belong to either \emph{`drinks'} or \emph{`dairy\_eggs\_fridge'}).
\end{itemize}

\begin{table*}[ht]
\centering
\caption{Performance comparison of evaluated models}
\label{tab:phase1-metrics}
\renewcommand{\arraystretch}{1}
\scriptsize{
\begin{tabular}{l c c c c c}
\toprule
\textbf{Model} & \textbf{Precision} & \textbf{Recall} & \textbf{F1} & \textbf{Accuracy} & \textbf{Balanced Accuracy} \\
\midrule
Claude 3.7 Sonnet & 0.907 & 0.902 & 0.905 & 0.902 & 0.773 \\
Claude 4 Sonnet   & 0.853 & 0.848 & 0.851 & 0.848 & 0.748 \\
Mixtral 8x7B      & 0.698 & 0.694 & 0.696 & 0.694 & 0.608 \\
Mistral 7B        & 0.604 & 0.596 & 0.600 & 0.596 & 0.492 \\ 
\bottomrule
\end{tabular}
}
\end{table*}

This dual-mode approach provided a more nuanced view of classification accuracy, distinguishing between
clear misclassification and contextually acceptable alternatives. 
Each model variant was analysed using the same metrics and tools as in Phase 1, including overall and balanced accuracy, precision, recall, F1 score, and per-category accuracy tables. 
In addition to accuracy metrics, each model variant was also compared on runtime, token counts, and estimated per-call costs to assess the trade-off
between improved accuracy and computational efficiency. 

Phase 2 analysis also quantified the proportion of
inherently ambiguous items (approx. 20 of the 389 total) that affected strict accuracy but were acceptable under lenient evaluation.
Manual verification was conducted for a sample of receipts to confirm that improvements were attributable to prompt refinements rather than incidental variation. Comparative tables and graphs summarised these results across all four variants to identify the most balanced configuration for production.

The same cost-control measures from Phase 1 were retained: all experiments were executed locally using only the Bedrock InvokeModel API, with results cached to avoid redundant inferences. The main cost difference in Phase 2 arose from longer prompt text and, in the few-shot variant, embedded examples, both of which increased input token counts. Output length remained stable across variants. As a result, input token usage became the primary cost driver, with the most tuned variant roughly doubling the per-call cost compared to the baseline. 
These findings were integrated into the accuracy-to-cost comparison to support model selection for deployment.

\section{\uppercase{Results}}
\label{sec:results}

\subsection{Phase 1: Findings}
\label{sec:results_phase1}

The analysis of the results produced within Phase 1 of the study demonstrated that the Claude 3.7 Sonnet foundational model offers the best balance, combining high accuracy with steady runtimes and perfectly aligned output lengths. 
Claude 4 Sonnet performs well but is slower on
average, so it does not improve the overall trade-off for this type of task. 
The open-weight models are generally faster and likely cheaper to operate, yet they were markedly less reliable, with frequent array-length mismatches and category-level hallucinations. 
Based on this, Claude 3.7 is the recommended candidate for Phase 1, advancing to Phase 2 for targeted refinements. 
In what follows, we discuss the results of Phase~1 analysis in detail.

\begin{table*}[ht]
\centering
\caption{Category-wise performance comparison across models}
\label{tab:phase1-metrics-cat}
\scriptsize{
\renewcommand{\arraystretch}{1}
\begin{tabular}{l c c c c c}
\toprule
\textbf{Category} & \textbf{Count} & \textbf{Claude 3.7} & \textbf{Claude 4} & \textbf{Mixtral 8x7B} & \textbf{Mistral 7B} \\
\midrule
\texttt{fresh\_produce} & 31 & 1.000 & 1.000 & 0.903 & 0.742 \\
\texttt{meat\_seafood} & 22 & 0.864 & 0.727 & 0.773 & 0.636 \\
\texttt{dairy\_eggs} & 30 & 0.867 & 0.867 & 0.800 & 0.700 \\
\texttt{pantry\_snacks} & 90 & 0.967 & 0.878 & 0.678 & 0.567 \\
\texttt{bakery} & 17 & 0.882 & 0.882 & 0.765 & 0.706 \\
\texttt{coffee\_tea} & 14 & 0.857 & 1.000 & 0.643 & 0.714 \\
\texttt{beverages} & 25 & 0.800 & 0.760 & 0.760 & 0.720 \\
\texttt{alcohol} & 5 & 0.800 & 0.000 & 0.400 & 0.400 \\
\texttt{eating\_out} & 61 & 0.984 & 1.000 & 0.820 & 0.492 \\
\texttt{health\_medicine} & 6 & 1.000 & 1.000 & 1.000 & 1.000 \\
\texttt{personal\_care\_beauty} & 28 & 1.000 & 0.929 & 0.214 & 0.643 \\
\texttt{home\_cleaning} & 21 & 0.857 & 0.619 & 0.429 & 0.333 \\
\texttt{pets} & 1 & 0.000 & 1.000 & 0.000 & 0.000 \\
\texttt{clothing\_footwear} & 10 & 0.900 & 0.900 & 0.800 & 0.900 \\
\texttt{electronics\_tech} & 1 & 1.000 & 1.000 & 1.000 & 1.000 \\
\texttt{stationery\_office} & 4 & 0.500 & 0.750 & 1.000 & 0.250 \\
\texttt{sports\_fitness} & 1 & 1.000 & 1.000 & 0.000 & 0.000 \\
\texttt{gifts\_occasions} & 2 & 1.000 & 1.000 & 0.000 & 0.000 \\
\texttt{subscriptions\_digital} & 1 & 1.000 & 1.000 & 1.000 & 0.000 \\
\texttt{professional\_services} & 6 & 0.333 & 0.167 & 0.833 & 0.333 \\
\texttt{utilities\_bills} & 5 & 0.600 & 0.000 & 0.600 & 0.600 \\
\texttt{travel\_holidays} & 1 & 0.000 & 0.000 & 0.000 & 0.000 \\
\texttt{other} & 7 & 0.571 & 0.714 & 0.571 & 0.571 \\
\bottomrule
\end{tabular}
}
\end{table*}

\subsubsection{Overall Accuracy Comparisons}
Model performance was evaluated by comparing each predicted category against the manually verified
\texttt{ground\_truth\_category} column in the Phase 1 dataset. 
A prediction was counted as correct only when it
exactly matched the corresponding ground-truth label for that item. 
Using these results, we calculated overall
accuracy, balanced accuracy, precision, recall, and F1 scores for each model.  
Table~\ref{tab:phase1-metrics} summarises these metrics.  
The results highlight the clear performance
separation between models:
Claude 3.7 provides generally stronger results, with Claude 4 being close behind, while Mixtral and Mistral are significantly behind.

The results indicate that the Claude models produce more consistent, correctly structured outputs, which align with the lower hallucination rates observed during evaluation. For production use, these findings suggest that Claude 3.7 offers the best balance between precision and recall, while open-weight models trade accuracy for speed and cost.

\subsubsection{Category-based Accuracy Comparisons} 

Category-based accuracy was calculated by comparing each model’s predicted label for every item against the
corresponding \texttt{ground\_truth\_category} value in the manually verified dataset. 
These calculations were performed automatically by the evaluator script, 
which grouped predictions by category and computed the proportion of correct classifications within each group. This approach highlights differences in model performance across both common and infrequent expense types while reducing the influence of dominant categories. 

Table~\ref{tab:phase1-metrics-cat} presents the results for all four models. Categories with larger sample sizes (such as \emph{`Pantry \& Snacks'} or \emph{`Eating Out'}) provide a stable basis for comparison, while those with a smaller number of samples  can vary sharply because even a single error can substantially alter the percentage.

Thus, the categories with larger sample sizes remain strong for the Claude models, particularly \emph{`Eating Out'}, \emph{`Fresh Produce'}, and \emph{`Pantry \& Snacks'}. 
Performance gaps widen in household and personal-care categories where item descriptions are less standardised, with the open-weight models showing more frequent misclassification or apparent category-level hallucinations. 
These errors typically occur between semantically adjacent categories rather than as random outputs, reflecting the natural ambiguity of many receipts.
Results for very small categories vary widely because single predictions can disproportionately affect percentages. \emph{`Pets'},
\emph{`Gifts \& Occasions'}, and 
\emph{`Subscriptions \& Digital Services'}  illustrate this sensitivity and
should be interpreted with caution. 

Overall, the pattern reinforces the need for balanced accuracy reporting and category-based analysis to offset category imbalance. 
Using these measures ensures each category
contributes equally to the overall performance score, preventing high-volume classes from dominating the evaluation and masking weaknesses in low-frequency categories. This makes it easier to isolate which specific categories require targeted improvements, such as refined prompts or additional training examples.  
In practice, Claude 3.7 demonstrated the highest stability across common categories, suggesting that targeted disambiguation rules or modest data augmentation could further improve performance in sparse classes during later phases.

\subsubsection{Output Array-length Accuracy Comparisons} 

Array-length consistency was evaluated by comparing the number of elements in each model’s JSON output
array with the number of detected items in the ground-truth data for every receipt. A mismatch occurred when these two counts differed, indicating that the model either produced extra category labels, omitted some items, or returned a truncated array. The mismatch rate is the proportion of receipts with at least one such discrepancy.  

Array-length consistency was perfect for the Claude models (mismatch rate 0.00\% for both Clause 3.7 and Claude 4), which always returned a JSON array matching the ground-truth item count, eliminating downstream parsing errors and category-level hallucinations. 
By contrast, both open-weight models produced occasional to frequent mismatches, with Mixtral~8x7B demonstrating a mismatch rate of approx.~1.9\% and
Mistral~7B exhibited the highest instability, including extra labels on single-item receipts, resulting in a mismatch rate of approx.~9.3\%. These discrepancies
indicate weaker adherence to schema constraints and increase the risk of hallucinated or truncated outputs in
production.

\subsubsection{Runtime Performance Comparisons} 

Table~\ref{tab:phase1-runtime} summarises the runtime performance results. 
These results demonstrate  that the open-weight models are generally faster, with Mixtral~8x7B
leading in mean latency, while Claude 3.7 is consistently quicker than Claude 4 within the Claude family.
However, tail behaviour matters: Mistral 7B exhibits extreme worst-case latency, and both open-weight models show higher variance that coincides with hallucinated or malformed outputs. In contrast, the Claude models are slower but steadier, trading a few hundred milliseconds for more reliable, schema-conformant responses. 

\begin{table}[ht]
\centering
\caption{Performance comparison of evaluated models}
\label{tab:phase1-runtime}
\renewcommand{\arraystretch}{1}
\scriptsize{
\begin{tabular}{l c c c  }
\toprule
\textbf{Model} & \textbf{Mean (ms)} & \textbf{Min (ms)} & \textbf{Max (ms)}   \\
\midrule
Claude 3.7 Sonnet & 1265 & 873 & 3078   \\
Claude 4 Sonnet   & 1433 & 938 & 2852 \\
Mixtral 8x7B      & 431 & 208 & 2144   \\
Mistral 7B        & 912 & 172 & 15351 \\ 
\bottomrule
\end{tabular}
}
\end{table}

\subsubsection{Token Usage and Estimated Cost Comparisons} 

Token accounting was available only for the Claude models, which consistently reported both input and output token counts. On average, each receipt classification required roughly 420 input tokens and 24 output tokens, resulting in an estimated cost of \$0.004 per call under Bedrock’s published pricing~\cite{AWSpricing}. This equates to approximately 250 inferences per dollar. 
Token usage between Claude 3.7 and Claude 4 was nearly identical, confirming that prompt and output lengths were well controlled across both models. 
Open-weight models such as Mixtral~8x7B and Mistral~7B did not provide token metrics, but given their smaller architecture and shorter prompts, their operational costs are expected to be lower, though this could not be quantified within Bedrock’s reporting framework.

Table~\ref{tab:phase1-token-claude}  presents the estimated token usage and the estimated cost per call for Claude models.\\
~

\begin{table}[ht]
\centering
\caption{Token usage and estimated cost per call by Claude Sonnet Models }
\label{tab:phase1-token-claude}
\renewcommand{\arraystretch}{1}
\scriptsize{
\begin{tabular}{l l l }
\toprule
\textbf{Metric} & \textbf{Claude 3.7} & \textbf{Claude 4} \\
\midrule 
average number of input tokens & 421 & 422 \\ 
min. number of input tokens & 336 & 337 \\ 
max. number of input tokens & 1083 & 1084 \\ 
average number of output tokens & 24 & 24 \\ 
min. number of output tokens& 6 & 6 \\ 
max. number of output tokens & 165 & 156 \\
\hline
average input cost/call (USD)& 0.003370 & 0.003380 \\
average output cost/call (USD)& 0.000580 & 0.000580 \\
average total cost/call (USD) & 0.003950 & 0.003960 \\
number of calls per USD 1 & 250      & 250      \\
\bottomrule
\end{tabular}
}
\end{table}

\subsection{Phase 2: Findings}
\label{sec:results_phase2}

The Variant 3 model (Claude 3.7 Sonnet baseline model, coupled with updated categories and rules,  zero-shot prompting), achieved
the highest strict and lenient accuracies, confirming that clear category definitions and simple disambiguation rules reduce overlap errors. The trade-off is cost: this variant roughly doubles per-call spend compared to the baseline due to longer prompts. If the budget is tighter, updated categories only (no-shot) offers a strong middle ground, delivering meaningful accuracy gains with modest cost and runtime increases. Few-shot
prompting increased the number of input tokens and cost without improving results beyond the rules-only
variant, so it is not recommended for production.

\subsubsection{Accuracy Analysis}

The strict evaluation shows clear gains from prompt refinements, see Table~\ref{tab:strict_evaluation_metrics}. Updating the categories alone yields a small lift in overall metrics and a larger improvement in balanced scores, indicating better handling of minority classes. 

Adding rules/heuristics delivers the strongest results, with overall precision/recall/F1 around 93\% and balanced metrics also improving over the baseline, suggesting that explicit disambiguation reduces overlap errors across adjacent categories.

By contrast, few-shot examples do not outperform the rules-only variant under strict scoring and add input tokens. This aligns with qualitative observations: many of the 20 (out of 389) inherently ambiguous items (e.g., `iced
coffee' as \emph{`drinks'} vs. \emph{`dairy\_eggs\_fridge'}) penalise strict metrics regardless of examples. Overall, the Variant 3 model achieves the best accuracy while avoiding unnecessary token growth from examples.

 The lenient evaluation shows clear gains from prompt refinements, see Table~\ref{tab:strict_evaluation_metrics}.  
 Under a lenient criterion (counting a prediction as correct if it matches either the primary label or a valid alternative), accuracy improves across all tuned variants. Variant 3 ranks highest, achieving approx. 95\% for precision, recall and F1 metrics for overall accuracy and approx. 85-86\% for balanced accuracy.
 
 The few-shot variant provided similar but slightly lower results for overall accuracy, while slightly improving for balanced accuracy. The lift is concentrated in previously ambiguous areas; for example, items such as iced coffee or hummus no longer penalise the model when mapped to a defensible alternative. In practice, this means only about 5\% of items are truly
incorrect under realistic user expectations.

\begin{table}[ht]
\centering
\caption{Strict Evaluation: Overall and Balanced Metrics (Claude 3.7 model)}
\label{tab:strict_evaluation_metrics}
\scriptsize{
\renewcommand{\arraystretch}{1}
\begin{tabular}{ll c c c}
\toprule
\textbf{Variant} & \textbf{Accuracy} & \textbf{Precision} & \textbf{Recall} & \textbf{F1} \\
\midrule
Variant 1 & Overall   & 90.70\% & 90.20\% & 90.50\% \\
Variant 1 & Balanced   & 81.40\% & 77.30\% & 77.20\% \\

Variant 2 & Overall  & 90.70\% & 90.50\% & 90.60\% \\
Variant 2 & Balanced   & 77.20\% & 82.20\% & 77.40\% \\

Variant 3 & Overall   & 93.30\% & 93.10\% & 93.20\% \\
Variant 3 & Balanced  & 80.80\% & 83.90\% & 80.80\% \\

Variant 4 & Overall   & 92.50\% & 92.30\% & 92.40\% \\
Variant 4 & Balanced  & 80.20\% & 83.20\% & 79.80\% \\
\bottomrule
\end{tabular}
}
\end{table}

\begin{table}[ht]
\centering
\caption{Lenient Evaluation:  Overall and Balanced Metrics (Claude 3.7)}
\label{tab:lenient_evaluation_metrics}
\scriptsize{
\renewcommand{\arraystretch}{1}
\begin{tabular}{ll c c c}
\toprule
\textbf{Variant} & \textbf{Accuracy} & \textbf{Precision} & \textbf{Recall} & \textbf{F1} \\
\midrule
Variant 2
& Overall 
& 91.80\% & 91.50\% & 91.60\% \\

Variant 2
& Balanced  
& 81.20\% & 82.90\% & 80.80\% \\

Variant 3
& Overall  
& 95.60\% & 95.40\% & 95.50\% \\

Variant 3 
& Balanced 
& 86.30\% & 85.80\% & 85.40\% \\

Variant 4
& Overall 
& 95.40\% & 95.10\% & 95.20\% \\

Variant 4
& Balanced  
& 90.40\% & 89.90\% & 89.60\% \\

\bottomrule
\end{tabular}
}
\end{table}

Category-based patterns are consistent with Phase~1, see Table~\ref{tab:lenient_per_category_accuracy}: strong stability in high-support classes (e.g.,
\emph{`fruits\_vegetables'}, \emph{`frozen'}, \emph{`eating\_out'}), large gains for \emph{`dairy\_eggs\_fridge} under extra rules, and continued weakness for drinks, where boundary definitions remain vague. Small-sample categories remain volatile and should be interpreted cautiously.

\subsubsection{Runtime Performance Comparison}

The baseline variant (Variant 1) is marginally faster on average than the tuned prompts, with small step-ups in mean latency as rules and few-shot examples are added, see Table~\ref{tab:phase2-runtime}. 
The differences remain modest, indicating that richer instructions do not materially affect runtime, though they slightly increase tail latency (the highest max for the few-shot
variant). 
This pattern is consistent with longer prompt text driving small overheads, while evaluation mode
(strict vs lenient) does not affect inference time.

\begin{table}[ht!]
\centering
\caption{Performance comparison of the  variants for the lenient evaluation}
\label{tab:phase2-runtime}
\renewcommand{\arraystretch}{1}
\scriptsize{
\begin{tabular}{l c c c  }
\toprule
\textbf{Variant} & \textbf{Mean (ms)} & \textbf{Min (ms)} & \textbf{Max (ms)}   \\
\midrule
Variant 1 & 1265 & 873 & 3078   \\
Variant 2 & 1359 & 861 & 3367 \\
Variant 3 & 1398 & 985 & 3364   \\
Varinat 4 & 1439 & 972 & 3739 \\ 
\bottomrule
\end{tabular}
}
\end{table}

\begin{table*}[ht!]
\centering
\caption{Lenient Evaluation: Category-based Accuracy by Variant}
\label{tab:lenient_per_category_accuracy}
\scriptsize{
\renewcommand{\arraystretch}{1}
\begin{tabular}{l c c c c}
\toprule
\textbf{Category} & \textbf{Count} & \textbf{Variant 2} & \textbf{Variant 3} & \textbf{Variant 4} \\
\midrule
\texttt{fruits\_vegetables} & 26 & 100.00\% & 100.00\% & 100.00\% \\
\texttt{meat\_seafood\_deli} & 17 & 100.00\% & 88.20\% & 94.10\% \\
\texttt{dairy\_eggs\_fridge} & 41 & 78.00\% & 100.00\% & 100.00\% \\
\texttt{frozen} & 19 & 100.00\% & 100.00\% & 100.00\% \\
\texttt{pantry\_snacks} & 73 & 98.60\% & 98.60\% & 95.90\% \\
\texttt{bakery} & 15 & 86.70\% & 86.70\% & 93.30\% \\
\texttt{coffee\_tea} & 14 & 100.00\% & 100.00\% & 100.00\% \\
\texttt{drinks} & 23 & 69.60\% & 56.50\% & 56.50\% \\
\texttt{liquor} & 5 & 80.00\% & 80.00\% & 80.00\% \\
\texttt{eating\_out} & 61 & 96.70\% & 100.00\% & 98.40\% \\
\texttt{health\_medicine} & 11 & 90.90\% & 90.90\% & 81.80\% \\
\texttt{personal\_care\_beauty} & 28 & 100.00\% & 89.30\% & 92.90\% \\
\texttt{cleaning\_maintenance} & 11 & 100.00\% & 90.90\% & 90.90\% \\
\texttt{baby\_maternity} & 0 & [N/A] & [N/A] & [N/A] \\
\texttt{pets} & 1 & 0.00\% & 100.00\% & 100.00\% \\
\texttt{clothing\_footwear} & 10 & 80.00\% & 80.00\% & 80.00\% \\
\texttt{electronics\_tech} & 1 & 100.00\% & 100.00\% & 100.00\% \\
\texttt{home\_lifestyle} & 15 & 93.30\% & 93.30\% & 86.70\% \\
\texttt{sports\_fitness} & 1 & 100.00\% & 100.00\% & 100.00\% \\
\texttt{gifts\_occasions} & 1 & 100.00\% & 100.00\% & 0.00\% \\
\texttt{entertainment} & 0 & [N/A] & [N/A] & [N/A] \\
\texttt{subscriptions\_digital} & 1 & 100.00\% & 100.00\% & 100.00\% \\
\texttt{professional\_services} & 1 & 100.00\% & 100.00\% & 100.00\% \\
\texttt{utilities\_bills} & 5 & 60.00\% & 60.00\% & 60.00\% \\
\texttt{transport\_fuel} & 0 & [N/A] & [N/A] & [N/A] \\
\texttt{travel\_holidays} & 1 & 0.00\% & 0.00\% & 0.00\% \\
\texttt{other} & 8 & 50.00\% & 87.50\% & 75.00\% \\
\bottomrule
\end{tabular}
}
\end{table*}

 \begin{table*}[ht!]
\centering 
\caption{Token usage and estimated cost per call by variants}
\label{tab:token_usage_claude37} 
\scriptsize{
\renewcommand{\arraystretch}{1}
\begin{tabular}{l c c c c}
\toprule
\textbf{Metric} & \textbf{Variant 1} & \textbf{Variant 2} & \textbf{Variant 3} & \textbf{Variant 4} \\
\midrule
average number of input tokens & 421  & 433  & 979   & 1212 \\
min. number of input tokens & 336  & 348  & 894   & 1127 \\
max. number of input tokens & 1083 & 1095 & 1641  & 1874 \\
average number of output tokens & 24   & 26   & 26.1  & 26.5 \\
min. number of output tokens & 6    & 6    & 6     & 6 \\ 
max. number of output tokens & 165  & 177  & 176   & 176 \\
\hline
average input cost/call (USD)  &  0.003370 &  0.003460 &  0.007830 &  0.009700 \\
average output cost/call (USD) &  0.000580 &  0.000620 &  0.000630 &  0.000640 \\
average total cost/call (USD)  &  0.003950 &  0.004160 &  0.008740 &  0.010670 \\
number of calls per USD 1       & 253        & 240        & 114        & 94 \\
\bottomrule
\end{tabular}
}
\end{table*}

\subsubsection{Token Usage and Estimated Cost Comparisons}

Input tokens increased with the refinements, mainly due to the added rules and few-shot examples, while output tokens remained nearly constant across all variants, see Table~\ref{tab:token_usage_claude37}. As a result, the estimated cost per call rose with each step of prompt refinement, with the rules-only variant roughly doubling the baseline cost and the rules plus few-shot variant
adding the most overhead. 
This reflects a clear trade-off: improvements in accuracy from structured prompts come at the cost of higher token usage, while output length and cost remain stable.

\subsection{{Limitations and Threats to validity}}
\label{sec:limitations}

Our experiments face several threats to validity. 
This study was designed as an initial qualitative assessment based on manual review. To reduce the potential for human error, all manual tasks were re-checked by at least one of the co-authors upon completion.  
 The dataset was collected by the team of authors from the same university, and, therefore, might lack the diversity and structured sampling needed for statistical diversity across vendors, receipt types, and image conditions.  
 However, even with these limitations, the findings provide valuable insights for shaping an in-depth study and developing an early prototype.  
%

\section{\uppercase{Conclusions}}
\label{sec:conclusions}

In this paper, we presented a systematic, cost-aware evaluation of large language models (LLMs) for receipt-item categorisation within a production-oriented classification framework. We compare four instruction-tuned models available through AWS Bedrock: Claude 3.7 Sonnet, Claude 4 Sonnet, Mixtral 8x7B Instruct, and Mistral 7B Instruct.  
This allowed us to answer \emph{RQ1. How effective can LLMs be for receipt items categorisation?} Our study demostrated, that even without any additional fine-tuning of prompting methods, the values of accuracy, precision, recall and F1 score might be approx. 0.9. 

Phase 1 of our study provided the answer to 
\emph{RQ2. What LLM might be especially appropriate for the task of receipt items categorisation?}

Across both evaluation phases, Claude 3.7 Sonnet consistently demonstrated the best balance of accuracy, consistency, and cost efficiency receipt-item classification task. 
In Phase 1 of our study, Claude 3.7 established a strong baseline, achieving high overall accuracy and zero array-length mismatches. The open-
weight models Mixtral 8x7B and Mistral 7B performed faster but were less reliable, frequently producing
inconsistent or hallucinated outputs. 

Phase 2 of our study provided the answer to 
\emph{RQ3. What might be the most appropriate prompting methods for the task of expense-items categorisation? } 
Phase 2 confirmed that refining the predefined categories and introducing rule-based heuristics further improved accuracy, while few-shot prompting added cost without meaningful gains.

Based on both phases of the study, we conclude that Claude 3.7 Sonnet, with the updated predefined categories and no-shot prompting, provides the optimal trade-off between accuracy and cost. This configuration consistently achieved high performance across both strict and lenient evaluations while keeping runtime and token usage within reasonable limits. It has therefore been selected as the receipt categoriser model, representing the best balance between practical accuracy and long-term cost optimisation.

\section*{{Acknowledgements}}

We would like to thank Shine Solutions for sponsoring this project under the research grant PRJ00002626, and especially Branko Minic and Adrian Zielonka for sharing their industry-based expertise and advice.

 \bibliographystyle{apalike}
{\small

}

\end{document}